\documentclass[a4paper, 10pt, conference]{ieeeconf}      

\IEEEoverridecommandlockouts                              

\usepackage{graphicx} 
\usepackage{booktabs}%

\usepackage{caption}

\usepackage{tikz}
\usetikzlibrary{shapes.geometric, arrows}
\tikzstyle{startstop} = [rectangle, rounded corners, minimum width=3cm, minimum height=1cm, text centered, text width=3cm, draw=black, fill=blue!0]
\tikzstyle{process} = [rectangle, minimum width=3cm, minimum height=1cm, text centered, text width=3cm, draw=black, fill=blue!0]
\tikzstyle{decision} = [diamond,aspect=2, text centered, text width=2cm, draw=black, fill=blue!0, minimum size=1em,node distance=2cm, inner sep=0pt]
\tikzstyle{arrow} = [thick,->,>=stealth]

\title{\LARGE \bf
Google Map Aided Visual Navigation for UAVs in GPS-denied Environment
}

\newcommand\blfootnote[1]{%
  \begingroup
  \renewcommand\thefootnote{}\footnote{#1}%
  \addtocounter{footnote}{-1}%
  \endgroup
}

\author{Mo Shan$^{1*}$, Fei Wang$^{1}$, Feng Lin$^{1}$, Zhi Gao$^{1}$, Ya Z. Tang$^{1}$, Ben M. Chen$^{2}$
\thanks{$^{1}$Temasek Laboratories, National University of Singapore, Singapore}%
\thanks{$^{2}$Department of Electrical $\&$ Computer Engineering, National University of Singapore, Singapore}%
\thanks{$^{*}$Email: shanmo@nus.edu.sg}
}

\begin{document}

\maketitle
\thispagestyle{empty}
\pagestyle{empty}

\begin{abstract}

We propose a framework for Google Map aided UAV navigation in GPS-denied environment. Geo-referenced navigation provides drift-free localization and does not require loop closures. The UAV position is initialized via correlation, which is simple and efficient. We then use optical flow to predict its position in subsequent frames. During pose tracking, we obtain inter-frame translation either by motion field or homography decomposition, and we use HOG features for registration on Google Map. We employ particle filter to conduct a coarse to fine search to localize the UAV. Offline test using aerial images collected by our quadrotor platform shows promising results as our approach eliminates the drift in dead-reckoning, and the small localization error indicates the superiority of our approach as a supplement to GPS.

\end{abstract}

\section{INTRODUCTION}

Navigation of unmanned aerial vehicles (UAVs) in GPS-denied environment becomes increasingly critical. Since the UAVs often take off and land in different positions, it may not revisit the same scene. Thus, it may be difficult to detect loop closures as in simultaneous localization and mapping (SLAM). In addition, pose estimation via the fusion of inertial measurement unit (IMU) and optical flow (OF) suffers from drift \cite{honegger2013open}. To address these issues, we propose a geo-referenced localization framework, which is reliable, drift-free, and does not require loop closures.

We leverage on image registration to provide an absolute position in Google Map. Although its accessibility is appealing, and some relevant works have been reported, this task is quite demanding. Variation in scale, orientation, and illumination poses a great challenge to register the image captured by the onboard camera to the map. Furthermore, the scene changes between the onboard frame and the map is obvious because Google Map is not updated constantly.

To address these challenges, we rely on gradient patterns and use Histograms of Oriented Gradients (HOG) \cite{dalal2005histograms} for image registration. To expedite the matching process, we employ particle filter (PF) to avoid sliding window search. For efficiency, the search is confined around the UAV location predicted by OF.

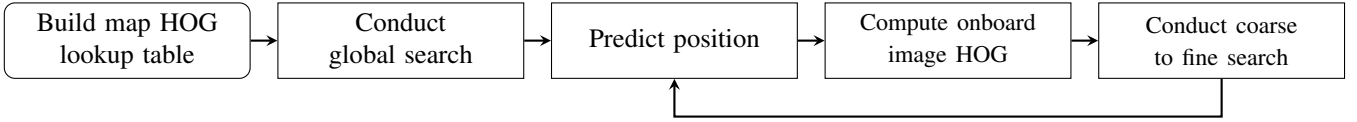
\begin{figure*}[t!]
  \centering
\begin{tikzpicture}[node distance=2.1cm]

\node (start) [startstop] {Build map HOG lookup table};
\node (pro2) [process, right of=start, xshift=1.5cm] {Conduct global search};
\draw [arrow] (start) -- (pro2);
\node (pro3) [process, right of=pro2, xshift=1.5cm] {Predict position};
\draw [arrow] (pro2) -- (pro3);
\node (pro4) [process, right of=pro3, xshift=1.5cm] {\small{Compute onboard image HOG}};
\draw [arrow] (pro3) -- (pro4);
\node (pro5) [process, right of=pro4, xshift=1.5cm] {\small{Conduct coarse to fine search}};
\draw [arrow] (pro4) -- (pro5);
\draw [arrow] (pro5.south) |- ++(0,-0.5) -| (pro3.south);

\end{tikzpicture}
  \caption{Overview of HOP. The HOG features for the map are computed offline. During onboard processing, we use global search to initialize the UAV position. Then for each frame, we track the pose by position prediction and image registration.}
  \label{fig:flowchart}
\end{figure*}

Since our approach combines HOG, OF and PF, we coin the terms and name it HOP. In short, our contributions are summarized as follows. Firstly, we present a simple yet effective navigation framework, which relies on correlation for initialization, HOG features to describe the images and PF to reduce the amount of comparisons. Secondly, we propose an OF based approach to compute inter-frame motion. To the best of our knowledge, this is the first time that low resolution Google Map and HOG are used for UAV navigation.

The rest of the paper is organized as follows: Section II presents a literature review; the detailed implementation of HOP is illustrated in Section III; Section IV contains experiments and analysis; following it are the conclusion and future research directions.

\section{Related work}

Various methods have been developed for UAV navigation to deal with GPS disruption. These works rely on different geographic information, including Geographic Information System (GIS), Google Earth, and Google Street View.

GIS data and its vector layers have been used to estimate UAV position in early works on geo-referencing. In \cite{patterson2011utilizing}, GIS data in the form of Ordnance Survey (OS) layers are divided into overlapping tiles, and inertial navigation system (INS) estimates which tile the UAV is located. The aerial image is rotated, scaled, and classified to form feature codes, which is compared with the OS data. Nevertheless, the training data may be inadequate to reflect the spectral signatures of all classes, especially under the presence of varying light conditions. Moreover, the classification, cross correlation and distance calculation are time consuming.

Images obtained via Google Earth have also been deployed for geo-referencing \cite{conte2008integrated, lindsten2010geo}. In \cite{conte2008integrated} the authors combine visual odometry and image registration to augment the navigation system. The Kalman filter is adopted to fuse the vision system with the INS. To compensate for the drift, an image registration technique realized by edge matching is developed. The registration is robust to change in scale, rotation and illumination to a certain extend. However, during the whole flight there are few successful matches. Therefore this method may not be suitable for long range flights.
Another approach \cite{lindsten2010geo} that relies on Google Earth images involves classification of the scene. UAV images are segmented into superpixels and then classified as grass, asphalt and house. Circular regions are selected to construct the class histograms, which are rotation invariant. However, discarding rotation gives rise to the classification uncertainty. Consequently sometimes the drift in position estimation is not successfully removed. Moreover, the matching accuracy is poor in large homogeneous regions. In addition, the training requires labeling the reference map manually.

Recently, several approaches \cite{majdik2013mav, le2014global} using Google Street View images for UAV localization have emerged. In \cite{majdik2013mav}, the aerial images are searched in the Google Street View database. To tackle with large viewpoint change, artificial views of the aerial images are generated and then Scale Invariant Feature Transform (SIFT) features are extracted. These features are compared against those extracted from ground images to find nearest neighbors. The outliers are removed via a histogram voting scheme and the good matches are verified by the Virtual Line Descriptor. Nevertheless, SIFT requires intensive computation and this approach does not use motion dynamics.

To summarize, our proposed approach mainly differs from the aforementioned ones in the following ways:
\begin{enumerate}
\item The easily accessible Google Map provides the geometric information for navigation, requiring less memory consumption compared with GIS and Google Street View.
\item The onboard sensors are utilized to obtain the rotation and scale of the frames, as well as the inter-frame translation.
\item It matches multi-modal images without feature detection and description. Instead, HOG is used holistically for image description and PF is employed to avoid sliding window search.
\end{enumerate}

\section{Geo-referenced navigation}

In this section the visual navigation framework will be investigated. An flowchart of HOP is given in Fig. \ref{fig:flowchart}, which provides an overview.

\subsection{Global localization}

After taking off, the UAV location is searched in the entire map for initialization. To avoid sliding window search, which is quite time consuming, we adopt the correlation filter proposed in \cite{bolme2010visual}. In Eq. \ref{eq:corr_filter}, F is the 2D Fourier transform of the input image, H is the transform of the filter, $\odot$ denotes element wise multiplication and * indicates complex conjugate. Because no training is available for detection, we correlate the current frame and the map. As a result, transforming G into the spatial domain gives a confidence map of the location.

\begin{equation} \label{eq:corr_filter}
G = F \odot H^{*}
\end{equation}

\begin{figure}[thpb]
  \centering
  \includegraphics[scale=0.195]{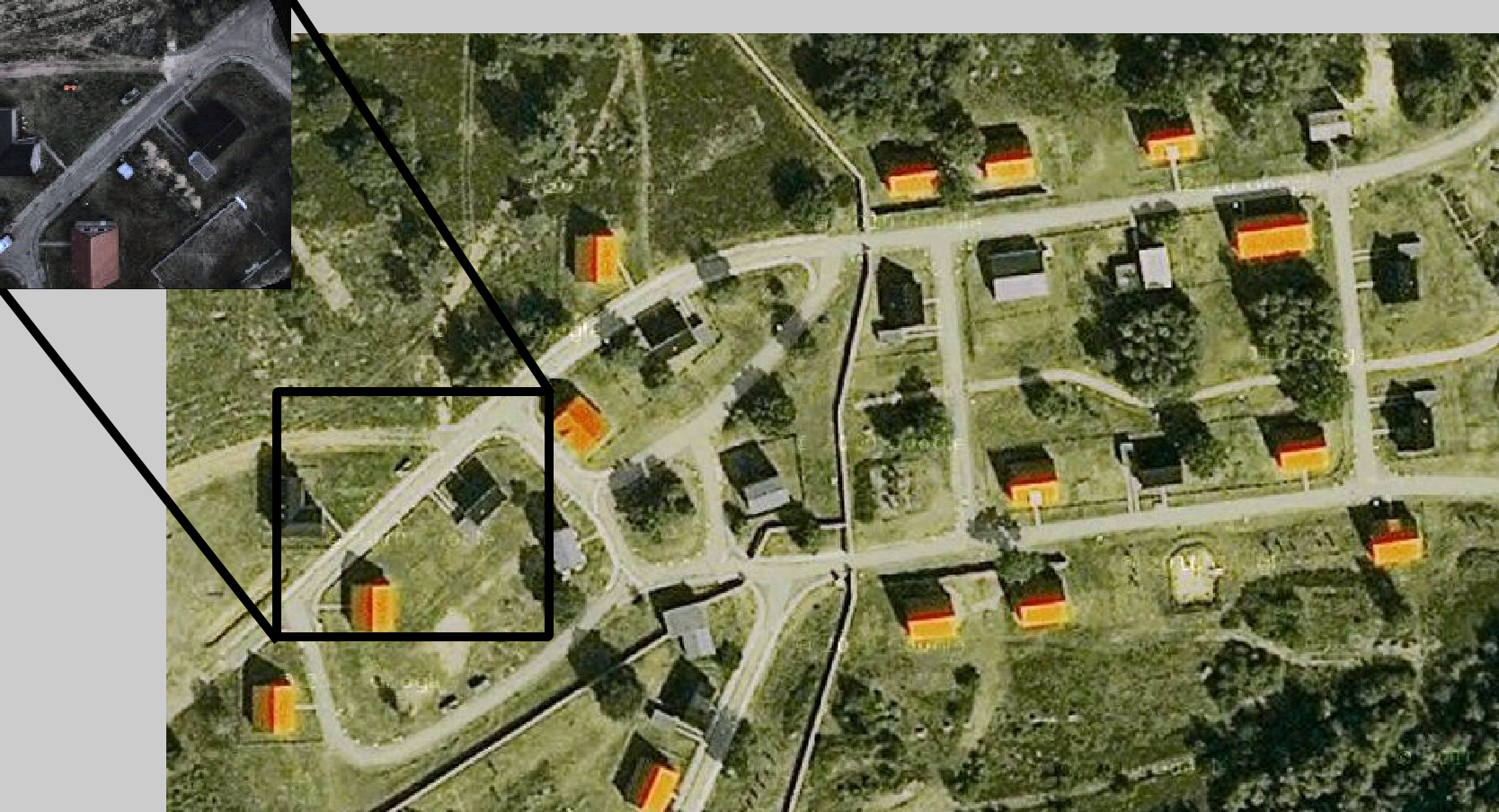}
  \caption{Onboard image at take off position, and its corresponding rectangular region in the map.}
  \label{fig:correlation_search}
\end{figure}

\begin{figure}[thpb]
  \centering
  \includegraphics[scale=0.165]{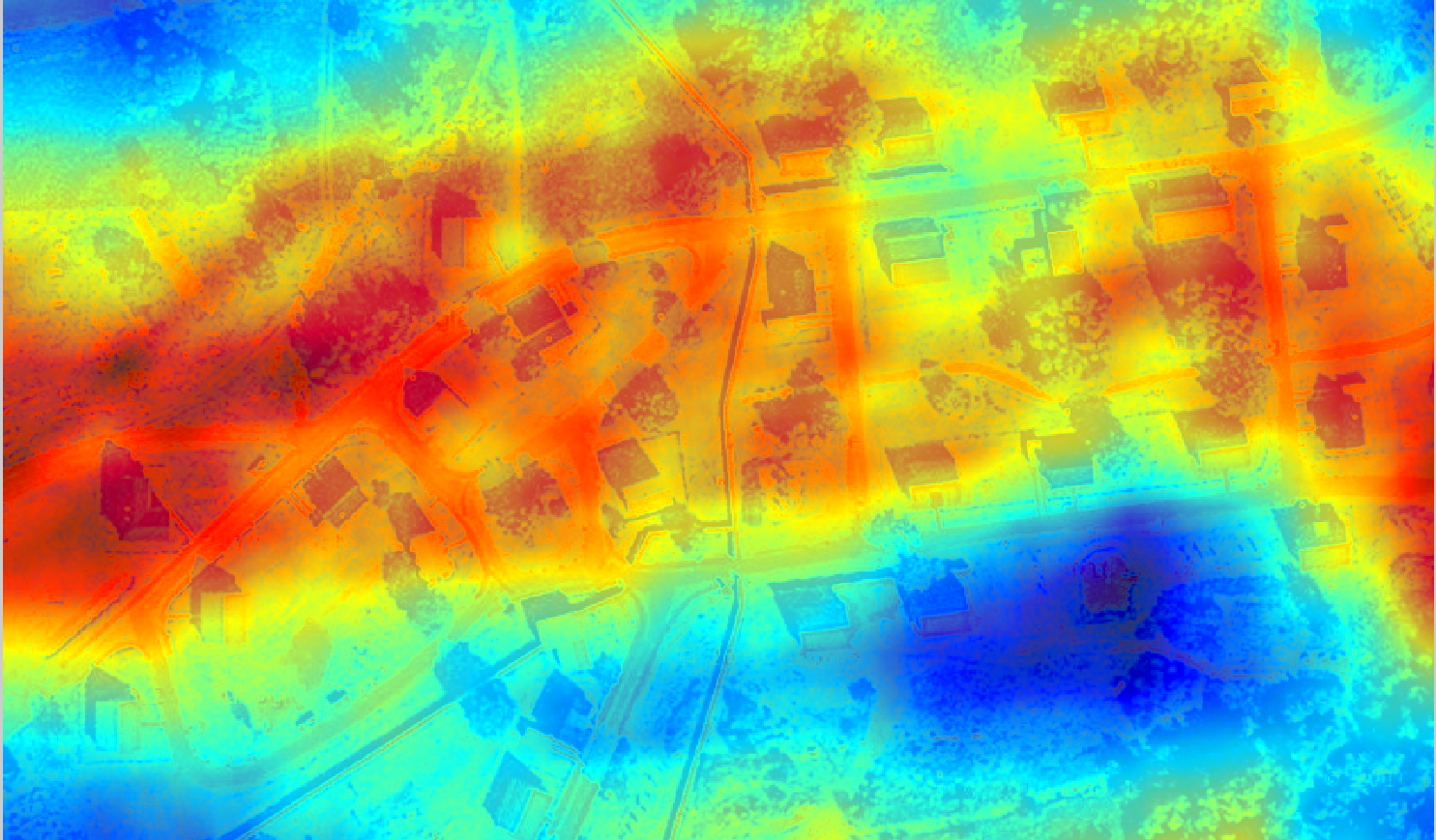}
  \caption{The confidence map of the frame. Red indicates high confidence while blue indicates low confidence. The black area represents the highest confidence, which suggests that the UAV is at take off position. Best viewed in color.}
  \label{fig:confidence_map}
\end{figure}

Take the onboard image displayed in Fig. \ref{fig:correlation_search} for example, it corresponds to the rectangular region on the map. Its confidence map is shown in Fig. \ref{fig:confidence_map}, from which it is evident that the correct location of the onboard image possesses the highest confidence. Although manual labeling may be more reliable, we still propose an autonomous global localization algorithm to make our framework complete.

\subsection{Pose tracking}

After initialization, the UAV position will be tracked based on local image registration. In this section, OF based motion estimation as well as HOG and PF based image registration will be introduced.

\subsubsection{Position prediction}

To narrow down the search, we make a rough guess and confine the matching around the predicted position by estimating the inter-frame motion. To obtain the motion, the points to be tracked are selected based on \cite{shi1994good}, and iterative Lucas-Kanade method with pyramids \cite{yves2000pyramidal} is used to construct the OF fields. The inter-frame translation could be derived from two approaches based on \cite{honegger2013open, zhao2012homography} respectively, both relying on supplementary information from the onboard avionic system and assuming the ground plane is flat.

\subsubsection*{Motion field}
For an interest point $P$, its coordinates $(x, y)$ in the camera frame and 3-D position $(X, Y, Z)^\textrm{\tiny{T}}$ are related by $x = f \cdot \frac{X}{Z}, y = f \cdot \frac{Y}{Z}$ according to projective projection, where $f$ is the focal length. Taking derivatives on both sides, we have $\dot{x} = v_x = f (\frac{\dot{X}}{Z} - \frac{X\dot{Z}}{Z^2}), \dot{y} = v_y = f (\frac{\dot{Y}}{Z} - \frac{Y\dot{Z}}{Z^2})$ where $v_x$, $v_y$ are the OF.

Let the camera motion be expressed as a translation, $\mathbf{T}=(T_x, T_y, T_z)^\textrm{\tiny{T}}$ and a rotation, $\Omega=(\omega_x, \omega_y, \omega_z)^\textrm{\tiny{T}}$, the velocity of the feature point $\mathbf{V}$ is defined by $\mathbf{V} = -\textbf{T} - \Omega \times P$, whose explicit form is (\ref{eq:sfm610}).
\begin{eqnarray}
\dot{X} &=& -T_x - \omega_y Z + \omega_z Y \nonumber \\
\dot{Y} &=& -T_y - \omega_z X + \omega_x Z \nonumber \\
\dot{Z} &=& -T_z - \omega_x Y + \omega_y X \label{eq:sfm610}
\end{eqnarray}

Therefore, $v_x$, $v_y$ are related to the motion by (\ref{eq:sfm612}).

\begin{eqnarray}
v_x - (-\omega_y f + \omega_z y + \frac{\omega_x xy}{f} - \frac{\omega_y x^2}{f}) = \frac{T_z x - T_x f}{Z} \nonumber \\
v_y - (-\omega_x f + \omega_z x + \frac{\omega_y xy}{f} - \frac{\omega_x y^2}{f}) = \frac{T_z y - T_y f}{Z} \label{eq:sfm612}
\end{eqnarray}

The rotational motion ($\omega_x$, $\omega_y$, $\omega_z$) are read from IMU and feature depth $Z$ is obtained by the barometer. Hence the terms on the left hand side are measurable. A linear equation set could be formulated for many feature points and the translation could be determined.

\subsubsection*{Homography decomposition}
Homography describes the relationship between co-planar feature points in two images, from which the motion dynamics could be derived according to (\ref{eq:h618}).
\begin{equation} \label{eq:h618}
\mathbf{H} = \mathbf{R} + \frac{1}{h}\mathbf{T}\mathbf{N}^\textrm{\tiny{T}}
\end{equation}
The $\mathbf{R}$ and $\mathbf{T}$ are the inter-frame rotation and translation, $\mathbf{N}$ is the normal vector of the ground plane, and $h$ is the altitude. $\mathbf{R}$, $\mathbf{N}$, $h$ are obtained from the onboard sensors and $\mathbf{T}$ can be calculated as (\ref{eq:h621}).
\begin{equation} \label{eq:h621}
\mathbf{T} = h(\mathbf{H}-\mathbf{R})\mathbf{N}
\end{equation}

\subsubsection{Image descriptor}

\begin{figure}[thpb]
  \centering
  \includegraphics[scale=0.45]{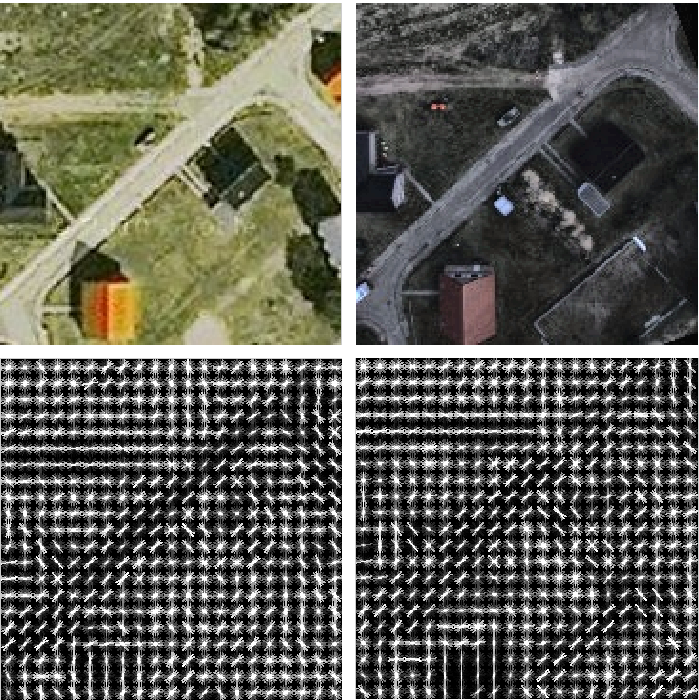}
  \caption{Visualization of HOG histograms. First row: subimage of reference map and onboard image. Second row: HOG glyph. The gradient patterns for houses and roads are quite similar in HOG glyph.}
  \label{fig:HOG_visual}
\end{figure}

Unlike object detection, no training is available in navigation. Therefore, we use HOG in a holistic manner as an image descriptor to encode the gradient information in multi-modal images.
The HOG glyph \cite{vondrick2013hoggles} is visualized in Fig. \ref{fig:HOG_visual}. It is evident that the gradient patterns remain similar even though the onboard image undergoes photometric variations compared with the map. In particular, the structures of road and house are clearly preserved.

During offline preparation phase, a lookup table is constructed to store the HOG features extracted at every pixel in the map. In this way, the HOG features for the map are retrieved from the table when registering images online to save computation time.

\subsubsection{Confined localization}

Comparison of HOG features is time consuming. Therefore, the traditional sliding window approach seems unfit as it demands more computational resources. Inspired by the tracking algorithms, we employ PF as in \cite{nummiaro2003adaptive} to estimate the true position. Furthermore, in order to reduce the number of particles, we confine our search in the vicinity of the predicted position, adopting a coarse to fine procedure.

\subsubsection*{Particle filter}

There are $N$ particles, and for each particle $p$, its properties include $\{x, y, H_{x}, H_{y}, w\}$, where $(x, y)$ specify the top left pixel of the particle, $(H_{x}, H_{y})$ is the size of the subimage covered by the particle and $w$ is the weight. The $(x, y)$ is generated around the predicted position, while $(H_{x}, H_{y})$ equals to the size of the onboard image.

The optimal estimation of the posterior is the mean state of the particles. Suppose each $p$ predicts a location $l$, then the estimated state is computed in Eq. \ref{eq:pf_mean}.

\begin{equation} \label{eq:pf_mean}
E(l) = \sum\limits_{i=1}^{N}w_{i}l_{i}
\end{equation}

Based on the predicted state $(x_{p}, y_{p})$ of where the UAV could be in the next frame, we calculate the likelihood that UAV location $(x_{c}, y_{c})$ is actually at this location.
After the particles are drawn, the subimages of the map located at the particles are compared with the current frame. To estimate the likelihood, we use Gaussian distribution to normalize these distance values based on Eq. \ref{eq:dist_norm}, where $d$ is the distance between the two images under comparison, $\sigma$ is the standard deviation, $\hat{w}$ is then normalized based on the sum of all weights to ensure that $w$ is in the range $[0, 1]$.

\begin{equation} \label{eq:dist_norm}
\hat{w} = \frac{1}{\sqrt{2\pi\sigma^2}} \exp(\frac{-d^2}{2{\sigma}^2})
\end{equation}

We do not use a dynamical model here to propagate the particles. Instead, we initialize the particles in every frame using OF estimation, similar to \cite{yao2010tracking}, for we have to conduct coarse to fine search and the particle number changes from frame to frame.

\subsubsection*{Coarse to fine search}

At the beginning, the particles are drawn around the take off position. Subsequently, OF provides translation between consecutive frames, and the predicted position is updated by accumulating the translation prior to image registration.

Around the predicted position, the search is conducted from coarse level to fine level to reduce the computational burden, similar to the coarse-to-fine procedure described in \cite{zhang2014fast}. For the coarse search, $N$ particles are drawn randomly in a rectangular area, whose width and height are both $s_{c}$, with a large search interval $\Delta_{c}$. The fine search, on the other hand, is carried out in an smaller area with size $s_{f}$ and search interval $\Delta_{f}$. Different from \cite{zhang2014fast}, HOP relies mainly on coarse search which is often quite accurate. If the minimum distance of coarse search is larger than a threshold $\tau_{d}$, then the match is considered invalid. Only when coarse search fails to produce valid match do we conduct fine search. Fine search still centers at the predicted position and the coarse search result is discarded.

When the minimum distances in both coarse and fine search are above the threshold $\tau_{d}$, indicating that image registration result is unreliable, the predicted position by OF is retained as the current location. If the motion is too large, the UAV may conduct global search for re-initialization.

\section{EXPERIMENT}

To evaluate the performance of HOP, we use the aerial images we have collected, which are displayed in the video accompanying the paper. The platform used to collect these images will be described, and then the pre-processing procedure as well as the parameter setting are elaborated. Next, we run experiments to analyse the effect of OF, and compare two outlier rejection schemes. We then compare HOP with visual odometry based on OF alone used as baseline.

\subsection{Setup}

\begin{figure}[thpb]
  \centering
  \includegraphics[scale=0.17]{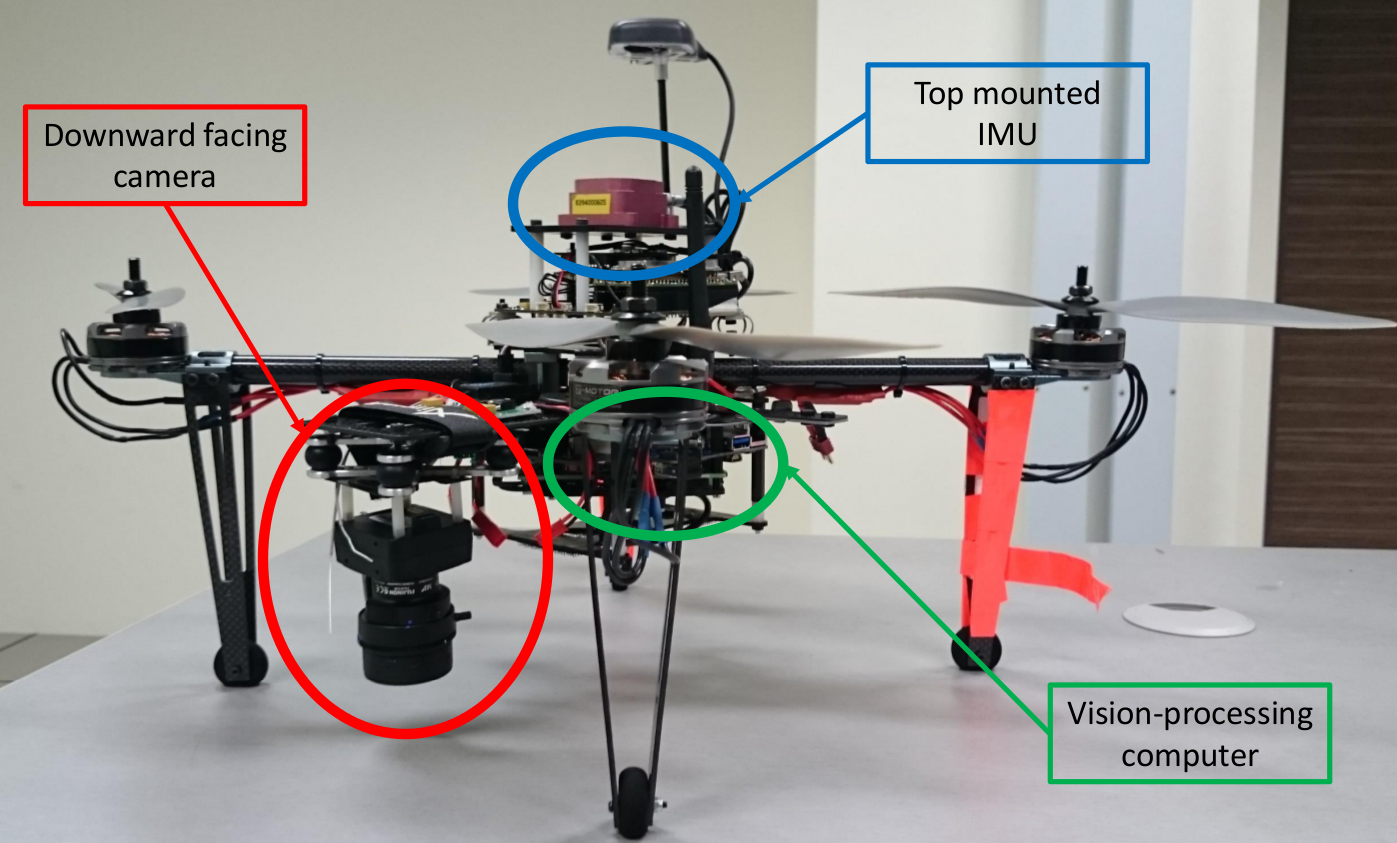}
  \caption{Quadrotor platform used for image capture, whose onboard sensors include IMU, Mastermind, and a camera.}
  \label{fig:quadrotor}
\end{figure}

In order to test the performance of the proposed algorithm, data is collected onboard using a quadrotor platform shown in Fig. \ref{fig:quadrotor}, whose dimension is 35 cm in height and 86 cm in diagonal width with a maximum take-off weight of 3 kg. Its onboard sensors include an IG-500N attitude and heading reference system (AHRS) from SBG Systems and a downward-facing PointGrey Chameleon camera. An Ascending Technologies Mastermind computer is used to decode and log data in a 5 Hz rate. The flight test is carried out at Oostdorp$^+$\blfootnote{$^+$Google Map location $[52.142815, 5.843196]$}, the Netherlands, and the onboard images and IMU data collected are the actual fly-off data for the final round of the 2014 International Micro Air Vehicle Competition (IMAV 2014). The quadrotor flies at about 80 m above the ground and sweeps overhead the whole Oostdorp village. The speed is about 2 m/s and the total flight duration is about 3 min.

\subsection{Preprocessing}

The reference with size $w\times h$ is the Google Map of Oostdorp village, which corresponds roughly to a $300\times 150$ m region. The resolution of the map is low, for about 3.15 pixels represent 1 m. The onboard frames are undistorted and then pre-processed to have the same orientation and scale as the reference map. First, the onboard frame is rotated by the yaw angle. Second, the frame is scaled to 3.15 pixels/m. The frames are cropped from the center with the same size $s_i\times s_i$.

\subsection{Parameters}

The most important parameters in HOP are $N$ and $s_{c}$. More $N$ increases the accuracy of the weighted center but demands more computational resources. Likewise, larger $s_{c}$ ensures the matching is robust to jitter while smaller $s_{c}$ reduces the time consumed. Hence, we trade off the robustness and efficiency when determining those parametric values.
Regarding the sensitivity of HOP to these parameters, it is found that $N$ should be larger than 40 to have sufficient particles to make a valid estimation. Meanwhile, $s_{c}$ should be larger than 35 to account for the inaccuracy arises from OF.

In our experiment, the varied parameters used are set as follows.
During preprocessing, $w\times h$ = $850\times500$, and $s_i$ = 180.
We use the HOG in OpenCV with cell size $32\times32$, block size $64\times64$, block stride $32\times32$.
For coarse to fine search, we set $N$ = 50, $s_{c} = 40$, $\Delta_{c} = 4$, $s_{f} = 20$, $\Delta_{f} = 1$, $\sigma$ = 0.01, $\tau_{d}$ = 0.75.

\subsection{Results}

\subsubsection{Effect of prediction}

\begin{figure}[thpb]
  \centering
  \includegraphics[scale=0.155]{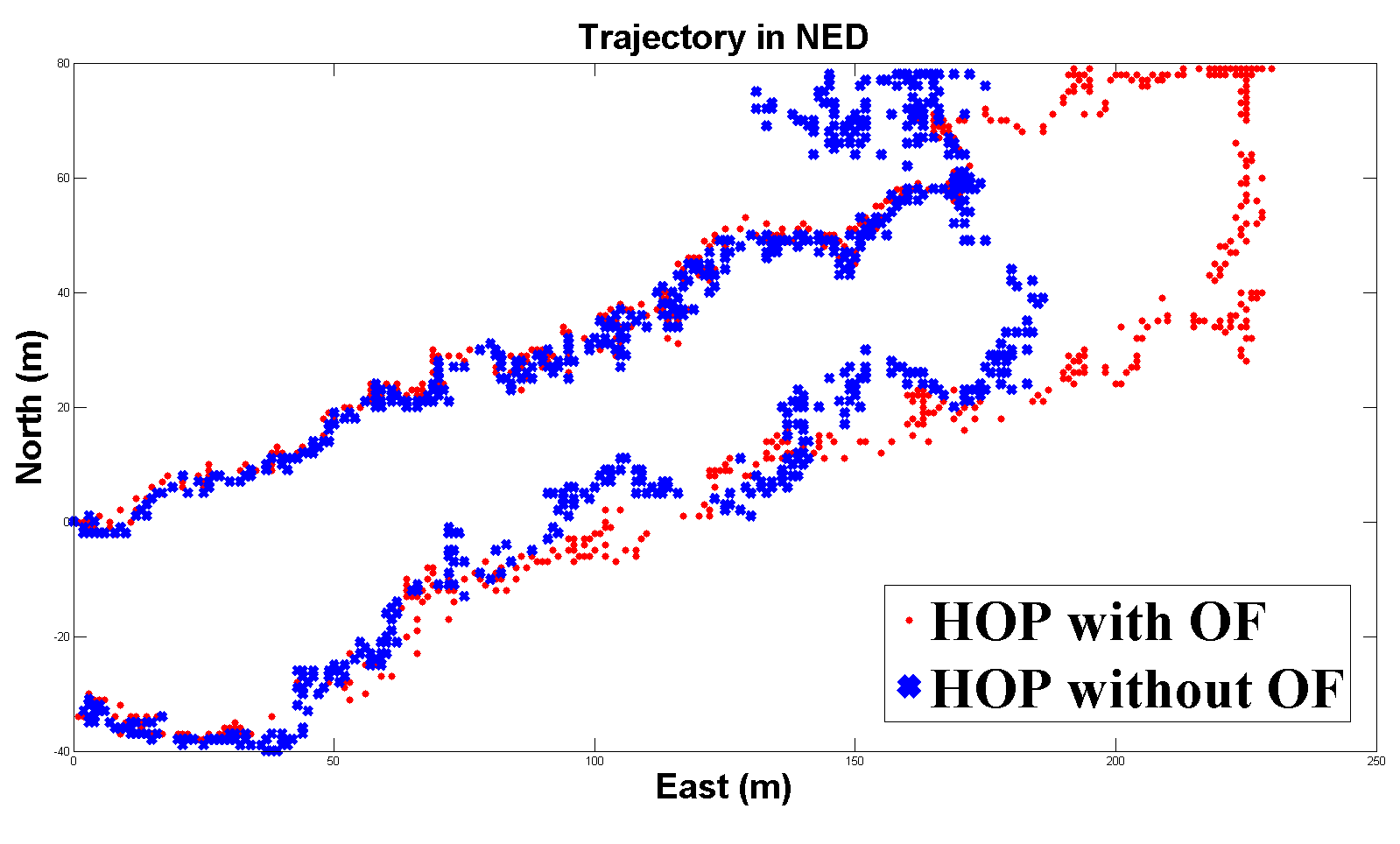}
  \caption{Comparison of HOP and HOP without OF. Red dots represent HOP, while blue crosses represent HOP without OF. Without OF, HOP fails when there is large motion or unreliable match, while using OF handles those issues effectively. Best viewed in color.}
  \label{fig:compare_of}
\end{figure}

We compare the effect of OF based position prediction as shown in Fig. \ref{fig:compare_of}. UAV localization fails without OF, especially when the motion is large. The localization is only resumed when the UAV flies close to a previously seen place. By contrast, using OF overcomes the large motion by moving the search region to the predicted position, and hence loop closure is unnecessary. 

\subsubsection{Rejection of outlier}

\begin{figure}[thpb]
  \centering
  \includegraphics[scale=0.155]{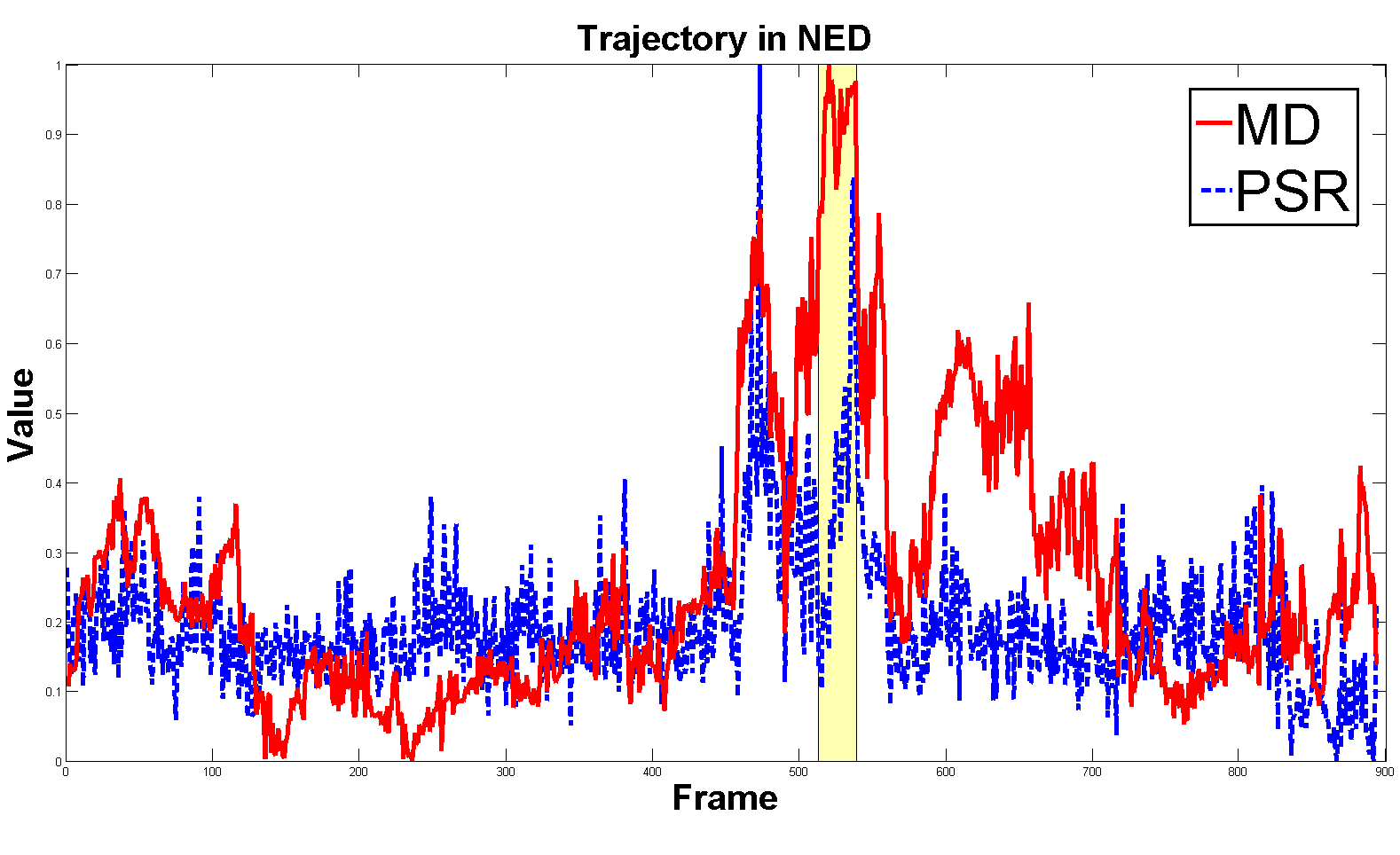}
  \caption{Comparison of outlier rejection methods. Red line depicts MD, whereas blue line depicts PSR. The yellow area highlighted corresponds to the frames with large illumination change, when the match becomes unreliable. MD is significantly higher for unreliable match in comparison to PSR. Best viewed in color.}
  \label{fig:outlier_rejection}
\end{figure}

\begin{figure*}[thpb]
  \centering
  \includegraphics[width=\textwidth,height=0.6\textwidth]{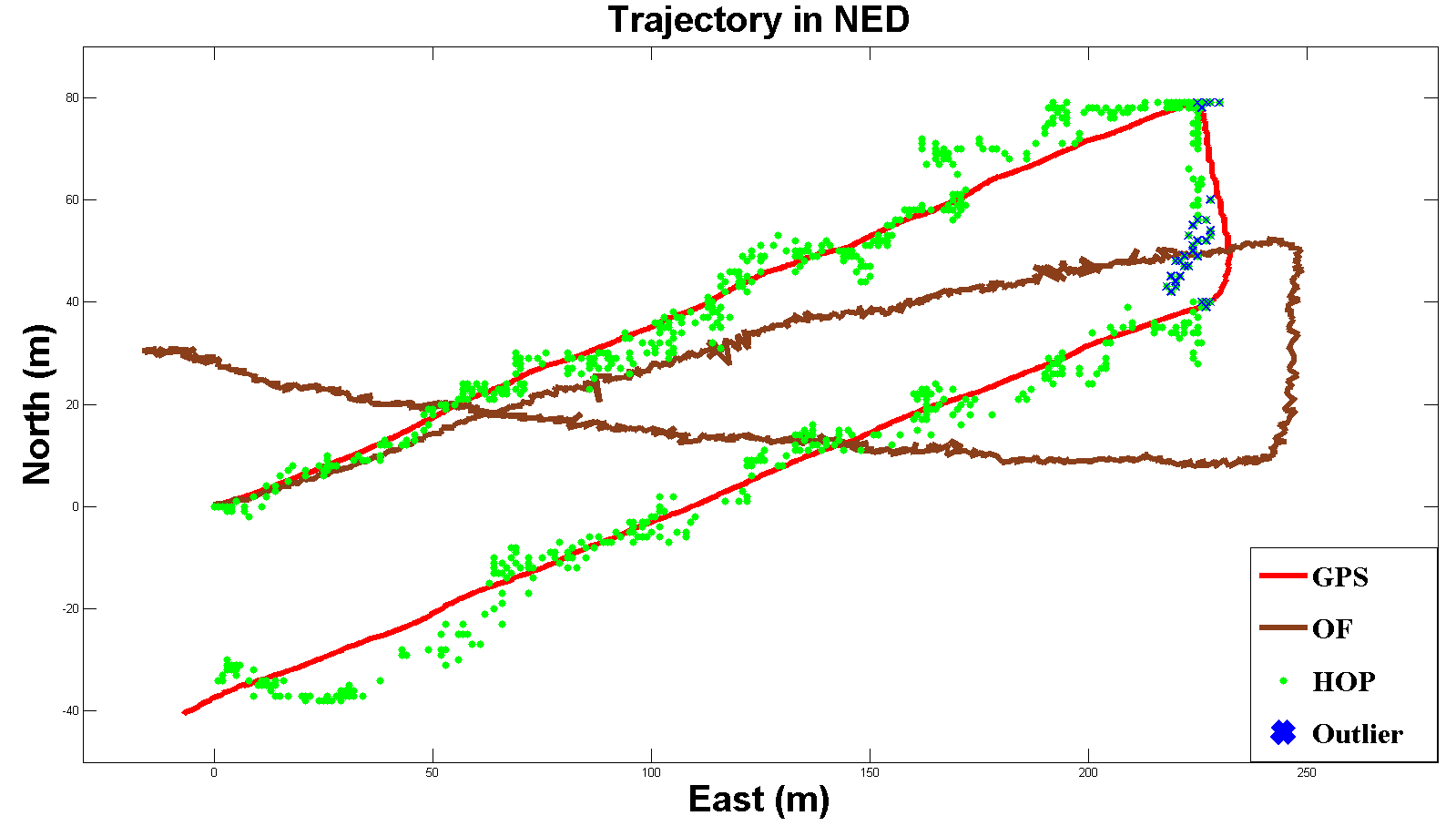}
  \caption{Path analysis comparing GPS (red line), OF (brown line), and HOP (green dots for reliable matches and blue crosses for outliers where OF is used). HOP performs markedly better than the OF baseline. Best viewed in color.}
  \label{fig:path_analysis}
\end{figure*}

We compare two methods to reject outliers, namely minimum distance (MD) as well as Peak to Sidelobe Ratio (PSR) defined in \cite{bolme2010visual}. PSR is computed according to Eq. \ref{eq:psr}, where $d_{min}$ is the minimum distance, and $\mu, \sigma$ are the mean and standard deviation of the distance for all particles in the search region, excluding a circle with 5 pixels radius around the minimum position.

\begin{equation} \label{eq:psr}
\Theta = \frac{d_{min} - \mu}{\sigma}
\end{equation}

The solid line in Fig. \ref{fig:outlier_rejection} is MD while the dashed line is PSR. Both MD and PSR are normalized to the range [0, 1]. MD peaks within the highlighted interval and attains large values, when the match becomes unreliable due to significant illumination change (refer to video). In contrast, PSR remains oscillating in that region. MD outperforms PSR in the sense that it indicates when the match is incorrect.

\subsubsection{Comparison with baseline}

The red line in Fig. \ref{fig:path_analysis} depicts the GPS ground truth, the brown line indicates the localisation from visual odometry based on OF alone as baseline. The green dots represent the HOP output and the blue crosses are outliers. The sequence is challenging for image registration for three reasons. Firstly, the Google Map is not up to date, and the trees and buildings are missing in some region. Secondly, the map image only has low resolution, which may reduce the amount of visible gradient patterns. Moreover, the scene undergoes large illumination change (refer to video).

As shown in Fig. \ref{fig:path_analysis}, HOP is both accurate and reliable, because it takes advantage of both the accuracy of HOG based localisation, and the reliability of OF based position prediction.

The dead-reckoning of OF gives poor results as the drift accumulates over time. On the other hand, the green dots follow the GPS closely, which corroborates the effectiveness of HOG based image registration.
In comparison to ground truth, the root mean square error (RMSE) of HOP is 6.773 m. The errors are quite small compared with a 169.188 m RMSE for the visual odometry based on OF alone. In fact, the localisation accuracy of HOP is comparable with GPS, whose RMSE is 3 m.

Furthermore, the position prediction step in HOP deals with unreliable match effectively as well. When there is obvious illumination change around the second turn, the HOG based match produces low similarity, and the predicted position is closer to the ground truth. Hence the match is discarded as outlier.

Image registration failure constitutes $7\%$ where position prediction is retained. The outliers mainly concentrate at two regions, where either there are few gradient patterns in the scene or has significant illumination change. We could design a flight path to avoid these homogeneous regions.
Moreover, the oscillation of HOP is sometimes significant, mainly due to wind and jitter of UAV. A gimbal could be used to mitigate the oscillation.

\subsection{Speed}

HOP is implemented in C++ using OpenCV. It is not optimized for efficiency and runs at 15.625 Hz on average for each frame on a Intel i7 3.40 GHz processor. The current update rate of HOP is sufficient for the position measurement, since its output will be fused with onboard INS at 50 Hz \cite{zhao2012homography}. Practically, the resulting trajectory is smooth as long as HOP is faster than 10 Hz.

\section{CONCLUSIONS}

This paper presents the first study of localization using HOG features in GPS-denied environment by registering aerial images and Google Map. The experiment using flight data shows that HOP could supplement GPS since its error is comparatively small. As the dataset is limited, our approach constitutes an initial benchmark, and we will make the onboard images and the reference map publicly available for research and comparison purposes.

For future research directions, we will install a gimbal to stabilize the camera against wind and vibration, and a thermal camera for navigation at night. Subsequently, we will perform more evaluation on challenging environments, including day and night conditions. We will also use HOP to give state feedback to an actual system.




\section*{ACKNOWLEDGMENT}
The authors would like to thank the members of NUS UAV Research Group for their kind support.

\bibliographystyle{IEEEtran}
\bibliography{IEEEabrv,mybibfile}

\end{document}